\documentclass{article} 
\usepackage{graphicx}
\usepackage{lmodern}
\usepackage{enumitem}
\usepackage{amsmath}
\usepackage{amssymb}
\newcommand{\bx} {\textbf{x}}
\newcommand{\by} {\textbf{y}}
\newcommand{\bu} {\textbf{u}}
\newcommand{\tv} {\textbf{$v_{thr}$}}
\usepackage[top=1in, bottom=1in, left=1in, right=1in]{geometry}

\title{Consistent Bounded-Asynchronous Parameter Servers for Distributed ML}

\author{
Jinliang Wei, Wei Dai, Abhimanu Kumar, \\ Xun Zheng, Qirong Ho and Eric P. Xing \\
School of Computer Science\\
Carnegie Mellon University\\
Pittsburgh, PA 15213 \\
\texttt{jinlianw@cs,wdai@cs,abhimank@andrew,}\\
\texttt{xunzheng@cs,qho@cs,epxing@cs.cmu.edu}
}

\begin{document}
\setitemize{labelindent=0.5cm,labelsep=0.5cm,leftmargin=*}

\maketitle

\begin{abstract}
In distributed ML applications, shared parameters are usually replicated among 
computing nodes to minimize network overhead. Therefore, proper consistency 
model must be carefully chosen to ensure algorithm's correctness and 
provide high throughput. Existing consistency models used in general-purpose 
databases and modern distributed ML systems are either too loose to guarantee 
correctness of the ML algorithms or too strict and thus fail to fully exploit the 
computing power of the underlying distributed system.

Many ML algorithms fall into the category of \emph{iterative convergent 
algorithms} which start from a randomly chosen initial point and converge to optima by repeating iteratively a set of procedures. We've found that many such algorithms are  to 
a bounded amount of inconsistency and still converge correctly. This 
property allows distributed ML to relax strict consistency models to improve 
system performance while theoretically guarantees algorithmic correctness. 
In this paper, we present several relaxed consistency models for asynchronous 
parallel computation and theoretically prove their algorithmic correctness. 
The proposed consistency models are implemented in a distributed parameter 
server and evaluated in the context of a popular ML application: topic modeling.

\end{abstract}

\section{Introduction}
In response to the rapidly increasing interests in big data analytics,
various system frameworks have been proposed to scale out ML algorithms to 
distributed systems, such as Pregel 
\cite{Malewicz:2010:PSL:1807167.1807184}, Piccolo 
\cite{Power_piccolo:building}, GraphLab 
\cite{Low:2012:DGF:2212351.2212354} and YahooLDA \cite{Ahmed:2012:SIL:2124295.2124312}. 
Amongst them, parameter server \cite{Smola:2010:APT:1920841.1920931, NIPS2013_4894, lazytables-hotos2013, Power_piccolo:building, Smola:2010:APT:1920841.1920931} 
is a widely used system architecture and programming abstraction that may support a broad range of ML algorithms.
 Parameter server can be conceptualized as a (usually distributed) key-value store that stores model 
parameters and supports concurrent read and write accesses from distributed 
clients \cite{NIPS2013_4894}. In order to minimize the network overhead of 
remote accesses, shared parameters are (partially) replicated on client nodes and accesses 
are serviced from local replicas. Thus proper level of consistency guarantees 
must be ensured. A desirable consistency model for parameter server must 
meet two requirements: 1) Correctness of the distributed algorithm can be 
theoretically proven; 2) Computing power of the system is fully utlized. The
consistency model effectively decouples the system implementations from ML algorithms and can be used to reason about the quality of the solution
(such as the rate of variance reduction).

Maintaining consistency among replicas is a classic problem in database research. 
Various consistency models have been used to provide different levels of 
guarantees for different applications \cite{Terry:2013:RDC:2534706.2500500}. 
However, directly applying them usually fails to meet the requirements of a 
parameter server. It is difficult or impossible to prove algorithm correctness
based on a naive eventual consistency model as it fails to bound the delay of 
seeing other clients' updates while one client keeps making progress and that 
may lead to divergence. Stronger consistency models such as sequential 
consistency and linearizability require serializable updates and that 
significantly restricts parallelism of the application.

Consistency models employed in modern distributed ML system tend to fall into
two extremes: either sequential consitency or no consistency guarantee at all. 
For example, Distributed GraphLab serializes read and write accesses to vertices
and edges by scheduling vertex programs according to a carefully colored graph 
or by locking \cite{Low:2012:DGF:2212351.2212354}. Altougth such a model 
guarantees the correctness of the algorithm, it may under-utilize the 
distributed system's computing power. At the other extreme, YahooLDA 
\cite{Ahmed:2012:SIL:2124295.2124312} employs a best-effort consistency model where the system makes best effort to delivery updates but does not make any guarantee. Although the system empirically achieves good performance for LDA, there is no theoretical guarantee that the distributed implementation of the algorithm is correct. It is also unclear whether such system with loose consistency can generalize to a broad range of ML algorithms. In fact, the system can potentially fail if stragglers present  or the network bandwith is saturated.

Recently, \cite{NIPS2013_4894} presented the Stale Synchrnous Parallel (SSP) 
model, which is a variant of bounded staleness consistency model -- a form of 
eventual consistency. Simimlar to Bulk Synchrnous Parallel (BSP), an execution 
of SSP consists of multiple iterations and each iteration is composed of a 
computation phase and a synchronization phase. Updates are sent out only
during the synchronization phase. However, in SSP, a client is allowed to go 
beyond other clients by at most $s$ iterations, where $s$ is a threshold set by the application. In SSP,
accesses to shared parameters are usually serviced by local replicas and 
network accesses only occur in case the local replica is more than $s$ 
iterations stale. SSP delivers high throughput as it reduces network 
communication cost. \cite{NIPS2013_4894} shows SSP is theoretically sound
by proving the convergence of stochastic gradient descent.

Asynchronous parallel model improves system performance over BSP because
1) it uses CPU and network bandwidth in parallel (e.g. sending out updates whenever 
network bandwidth is available); 2) it does not enforce a synchronization 
barrier. Also, since the system makes best effort to send out updates 
(instead of waiting for synchoronization barrer), clients are more likely to
compute with fresh data. Thus it may bring algorithmic benefits. It is more 
difficult to maintain consistency in an asynchronous system as communcation may
happen anytime.

We have found that many ML algorithms are sufficiently robust to a bounded amount of inconsistency and thus admits consistency models weaker than serializability. By relaxing the consistency guarantees properly, the system may gain significant
throughput improvements. In this paper, we present a few high-throughput consistency models 
that takes advantange of the robustness of ML algorithms. As shown in \cite{NIPS2013_4894},
even though relaxed consistency improves system throughput, it may result in reduced algorithmic progress per-iteration (e.g., smaller covergence rate). All our proposed consistency models allows application developers to tune the strictness
of the consistency model to achieve the sweet spot. We present the following consistency models:

\textbf{Clock-bounded Asynchronous Parallel (CAP)}: We apply the concept of ``clock-bounded"
consistency of SSP to asynchronous parameter server. Unlike SSP where updates are sent out only during 
the synchrnonization phase, CAP propgates updates whenever the netowrk bandwidth is available. Similar 
to SSP, CAP guarantees bounded staleness - a client must see all updates older than certain timestamp.

\textbf{Value-bounded Asynchronous Parallel (VAP)} This consistency model guarantees a bound on the
absolute deviation between any two parameter replicas.

\textbf{Clock-Value-bounded Asnychronous Parallel (CVAP)} This model combines CAP and VAP to provide
stronger consistency guarantee. With such a guarantee, the solution's quality (e.g. asymptotic variance) can be assessed. 

\section{Consistency Models}
\label{sec:consistency-models}
In this section, we present the formal definitions of CAP, VAP and CVAP. Consisder a collection of $P$ workers 
which share access to a set of parameters.
A worker writes to a parameter $\theta$ by applying an update in the form of an associate and commutative 
operation $\theta \leftarrow \theta + \delta $. The asynchronous parameter server propogates out the
update at some point of time and the worker usually proceeds without waiting but may block occasionally to
maintain consistency. 
Thus the parameter server may accumulate a set of updates generated by a worker which are
not yet synchronized acroos all workers. All our consistency models ensure read-my-writes consistency.
That is, a worker sees all its previous writes whether or not the updates are synchronized with other workers. Intuitively, read-my-write consistency is desirable as it allows a worker to proceed with more fresh
parameter. Our consistency models also ensure FIFO consistency \cite{Lipton:CS-TR-180-88} - updates from 
a single worker are seen by all other workers in the order in which they are issued. Intuitively, this 
consistency guarantee ensures that the updates are handled as fairly as the network ordering and prevents a worker from being biased by a particular subset of updates from another worker.


\subsection{Clock-bounded Asynchronous Parallel (CAP) }

Informally, Clock-bounded Asynchronous Parallel (CAP) ensures all workers are making sufficient  progress forward,
otherwise, faster workers are blocked to wait for the slow ones. Progress of a worker is 
represented by ``clock", which is an integer which starts from 0 and is incremented at regular intervals. 
Updates generated between clock $[c - 1, c]$ are timestamped with $c$. The consistency model guarantees that a worker 
with clock $c$ sees all other clients' updates in the range of $[0, c - s - 1]$, where $s$ is a user-defined threshold.

We omit the proof of correctness for CAP as the analysis in \cite{NIPS2013_4894} applies to CAP as well.

\subsection{Value-bounded Asynchronous Parallel (VAP) }
\label{sec:vap}

\begin{figure}
  \centering

  \includegraphics[width=0.5\textwidth]{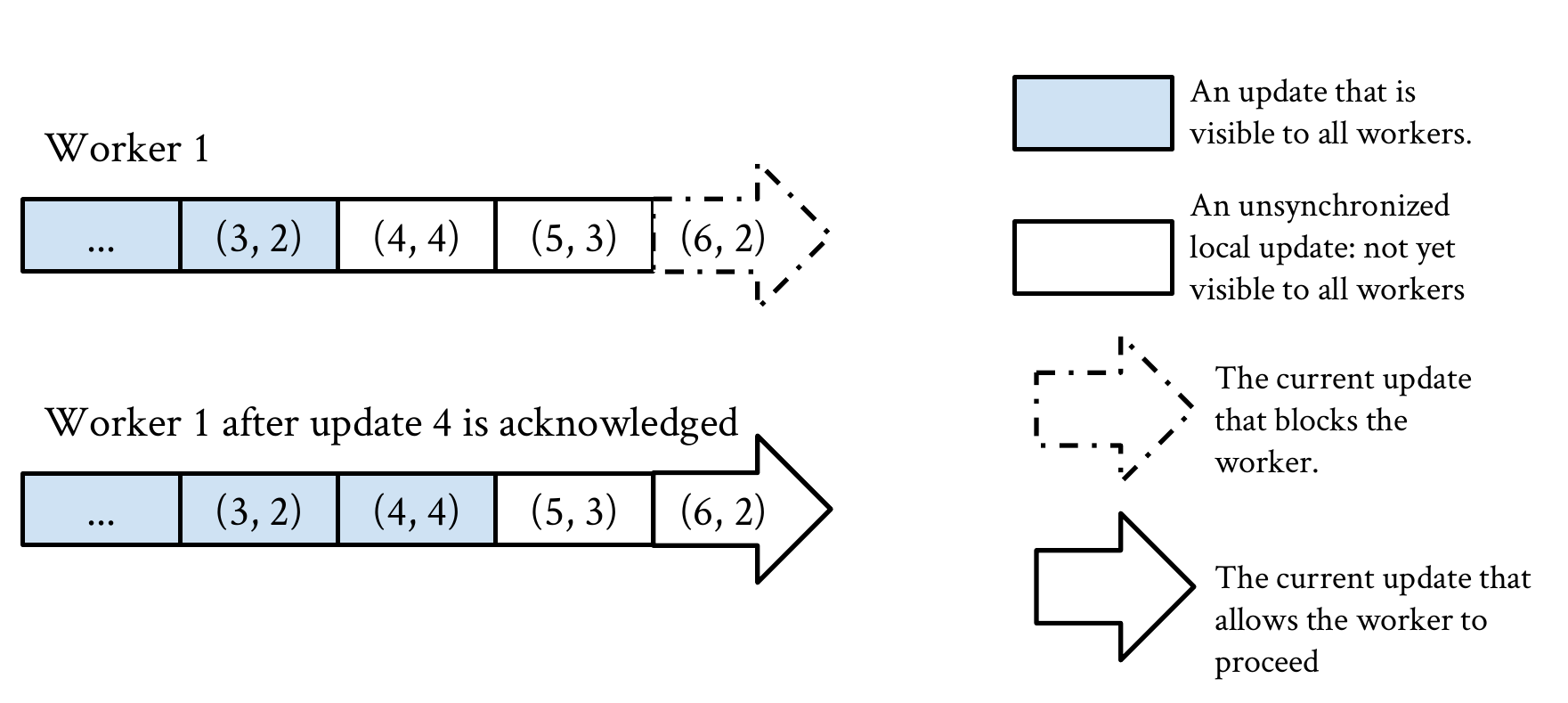}
    
    \caption{\small An illustration of VAP. The arrow represents a series of updates that a worker applies to a parameter.
        An update is representd by a pair of numbers $(i, j)$ where $i$ is the update's sequence number and $j$ is the value of the update. 
        In this example, the value bound is set to 8. Applying the 6th update blocks
        the worker as it would bring the accumulated unsynchronized updates beyond 8. After the 4th update becomes visible to all workers,
        the worker may procee with update $(6, 2)$.}
        \label{fig:weak_vap}
\end{figure}

Since the asynchronous parameter server does not block workers when propagating out updates, a worker
may accumulate a set of updates that are only visiable to itself.
We refer this set of updates as \emph{unsynchronized local updates}. As the update operation is associative and commutative, updates may be aggregated by summing them up.
The Value-bounded Asynchronous Parallel (VAP) model guarantees that for any worker the accumulated sum s 
of \emph{unsynchronized local updates} of any parameter is less than $\tv$ where $\tv$ is a user-defined threshold. When a worker attempts 
to apply an update that makes the accumulated sum exceed the threshold $\tv$, the worker is blocked until the system has made a sufficient number of its updates visible to all workers. The VAP model is illustrated by Figure~\ref{fig:weak_vap}.

We should note that the VAP model described above still allows two workers to see two very different set of updates. For any 
pair of workers, say A and B, even though VAP restricts how they see updates generated within this pair, it makes no guarantees about seeing updates that are generated by other peer workers. For example, it can happen that worker A has seen one update from each other worker, but B hasn't seen any of them. Thus, VAP only provides a very loose bound on how two workers may read different values: for any two workers A and B, let $\theta_A$ and $\theta_B$ be the sum over all updates seen by A and B respectively. Then, the absolute difference between $\theta_A$ and $\theta_B$, $|\theta_A - \theta_B|$, is upper bounded by $max(u, \tv) \times P$, where $u$ is an upper bound on the magnitude of any update, $\tv$ is the aforementioned user-defined threshold, and $P$ is the num

We may provide a stronger consistency guarantee by restricting how workers see other workers' updates. We refer to an update as a \emph{half-synchronized update} if the update is seen by at least other one worker (that did not generate the update), but has not yet been seen by all other workers. In addition to the guarantees provided by weak VAP, the strong VAP model guarantees that for any parameter, the total magnitude of all \emph{half-synchronized updates} is bounded by $max(u, \tv)$, where $u$ and $\tv$ are as defined before. Thus, strong VAP guarantees that for any two workers $A$ and $B$, $|\theta_A - \theta_B|$ is upper bounded by $2*max(u, \tv)$. Note that this is independent of $P$, the number of workers.

\subsection{Clock-Value-bounded Asynchronous Parallel (CVAP) }

CAP may be combined with VAP to provide stronger consistency guarantees. The idea is that CVAP ensure all workers make enough
progress but bounds the absolute difference between replicas. CVAP provides the consistency guarantees of both CAP and VAP. 
As VAP has a strong and a weak version, there are two versions of CVAP correspondingly. As we shown in Section~\ref{sec:theoretical-analysis}, CVAP allows application developers to guarantee a certain level of solution quality.

\section{Theoretical Analysis}
\label{sec:theoretical-analysis}
We choose the Stochastic Gradient Descent algorithm as our example application and
prove its convergence under VAP. Our proof utilizes the techniques developed in \cite{NIPS2013_4894}.
As in \cite{NIPS2013_4894}, the VAP model supports operations $\textbf{x} \leftarrow \textbf{x} \oplus (z\cdot \textbf{y})$, where
$\textbf{x},\textbf{y}$ are members of a ring with an abelian operator $\oplus$ (such as addition),
and a multiplication operator $\cdot$ such that $z\cdot \textbf{y} = \textbf{y}^\prime$ where $\textbf{y}^\prime$ is also in the ring. We shall informally refer to $\bx$ as the ``system state",
$\bu = z\by$ as an ``update", and to the operation $\bx \leftarrow \bx + \bu$ as ``writing an update".

As defined in  section~\ref{sec:vap} the updates 
$\bu$ are accumulated and are propagated to server when they
are greater than $\tv$. We call these propagation times as $t_p$.
We define $\bu_{p,t_p}$ as the accumulated update 
written by worker $p$ at propagation time $t_p$ through the write operation $\bx \leftarrow \bx + \bu_{p,t_p}$. 
The updates $\bu_{p,t_p}$ are
a function of the system state $\bx$, and under the VAP model, different workers will ``see" different, noisy versions of the true
state $\bx$. $\tilde{\bx}_{p,t_p}$ is the noisy state read by worker $p$ at time $t_p$, implying that $\bu_{p,t_p} = G(\tilde{\bx}_{p,t_p})$
for some function $G$. 
Formally, in VAP $\tilde{\bx}_{p,t_p}$ can take:

\paragraph{ Value Bounded Staleness:}
Fix a staleness $s$. Then, the noisy state $\tilde{\bx}_{p,t_p}$
is equal to
{\small
\begin{align}
\tilde{\bx}_{p,t_p} &= \bx_0
+ \underset{\text{guaranteed pre-propagation updates}}{\underbrace{\left[ \sum_{p^\prime=1}^{P} \sum_{t = 1}^{t_{p^\prime} - 1}
\bu_{p^\prime,t} \right]}}
+ \underset{\text{guaranteed read-my-writes updates}}{\underbrace{
\bu_{p,t_p} 
}}
+ \underset{\text{best-effort in-propagation updates}}{\underbrace{\left[ \sum_{(p^\prime,t_{p^\prime})\in \mathcal{S}_{p,t_p}} \bu_{p^\prime,t_{p^\prime}} \right]}},
\label{eq:ssp_cond}
\end{align}}

where $\mathcal{S}_{p,t_p} \subseteq \mathcal{W}_{p,t} = ([1,P] \setminus \{p\}) \times\{t_1,t_2,...,t_P\}$
is some subset of the updates $\mathbf{u}$ written in between propagations $t_p-1$ and $t_p$ ``window" and does not include updates from worker $p$.
In other words, the noisy state $\tilde{\mathbf{x}}_{p,t_p}$ consists of three parts:
\begin{enumerate}
\setlength{\itemsep}{0.6pt}
\setlength{\parskip}{1pt}
\item
Guaranteed ``pre-propagation" updates from begining to $t_{p^\prime}-1$, for every worker $p^\prime$.
\item
Guaranteed ``read-my-writes" set $\bu_{p,t_p}$ that covers all ``in-window" updates made by the querying worker $p$.
\item
Best-effort ``in-window" updates $\mathcal{S}_{p,t_p}$ 
(not counting updates from worker $p$). 

\end{enumerate}

As with \cite{NIPS2013_4894}, VAP also generalizes the Bulk Synchronous Parallel (BSP) model:
\paragraph{BSP Lemma:}
Under zero staleness VAP reduces to BSP. \textbf{Proof:} $\tilde{\bx}_{p,t_p}$ exactly consists of all updates until time $t_p$. $\square$

We now define a reference sequence of states $\bx_t$, informally referred to as the ``true" sequence :
\begin{align*}
\bx_t &= \bx_0 + \sum_{t^\prime=0}^{t} \bu_{t^\prime}, \qquad\qquad
\text{where}\quad \bu_t := \bu_{\text{$t$ mod $P$},\lfloor t/P \rfloor}.
\end{align*}
In other words, we sum updates by first looping over workers $(t$ mod $P)$, then over time-intervals $\lfloor t/P \rfloor$.
Now, let us use VAP to bound the difference between the ``true" sequence $\bx_t$ and the noisy views $\tilde{\bx}_{p,t_p}$:

\vspace{-0.2cm}
\paragraph{Lemma 1:}
Assume $s\ge 1$, and let $\tilde{\bx}_t := \tilde{\bx}_{\text{$t$ mod $P$},\lfloor t/P \rfloor}$, so that
\begin{align}
\tilde{\bx}_t &= \bx_t + \Delta_t, \label{eq:lemma1}
\end{align}
where $\Delta_t$ is the the difference (i.e. error) between $\tilde{\bx}_t$ and $\bx_t$.
We claim that $\Vert\Delta_t \Vert \le 2\tv\sqrt{K}(P-1)$, where $\Vert \cdot \Vert$ is the $\ell_2$ norm,
and $K$ is the dimension of $\bx$.
{\bf Proof}: As argued earlier, strong VAP implies $\Vert\Delta_t \Vert_\infty \le 2\tv(P-1)$. The result immediately follows since
$\Vert y \Vert \le \sqrt{K}\Vert y \Vert_\infty$ for all $y$.

\paragraph{Theorem 1 (SGD under VAP):}
Suppose we want to find the minimizer $\bx^*$ of a convex function $f(\bx) = \frac{1}{T} \sum_{t=1}^T f_{t}(\bx)$,
via gradient descent on one component $\nabla f_t$ at a time. We assume the components $f_t$ are also convex.
Let the updates $\bu_t := -\eta_t \nabla f_t (\tilde{\bx}_t)$, with decreasing step size $\eta_t = \frac{\sigma}{\sqrt{t}}$.
Also let the VAP threshold $\tv$ decrease according to $v_t = \frac{\delta}{\sqrt{t}}$.
Then, under suitable conditions ($f_t$ are $L$-Lipschitz and the distance between two points $D(x\Vert x^\prime) \le F^2$),
\begin{align*}
R[\mathbf{X}] &:= \sum_{t=1}^{T} \left[ f_t(\tilde{\bx}_t) - f(\bx^*) \right]  \le \sigma L^{2}\sqrt{T}+F^{2}\frac{\sqrt{T}}{\sigma}
+ 4 \delta LP\sqrt{KT}
\end{align*}
Dividing both sides by $T$, we see that VAP converges at rate
$\mathcal{O}\left(\frac{1}{\sqrt{T}}\right)$ when all other quantities are fixed.

\subsection{Proof of Theorem 1}
We follow the proof of \cite{NIPS2013_4894}.
Define $D\left(x\Vert x^{\prime}\right):=\frac{1}{2}\left\Vert x-x^{\prime}\right\Vert ^{2}$, where $\Vert \cdot \Vert$
is the $\ell_2$ norm.

\paragraph{Proof:}
We have ,
\begin{eqnarray*}
R\left[X\right] & := & \sum_{t=1}^{T}f_{t}\left(\tilde{x}_{t}\right)-f_{t}\left(x^{*}\right)\\
 & \le & \sum_{t=1}^{T}\left\langle \nabla f_{t}\left(\tilde{x}_{t}\right),\tilde{x}_{t}-x^{*}\right\rangle \qquad\text{(\ensuremath{f_{t}}\,\ are convex)}\\
 & = & \sum_{t=1}^{T}\left\langle \tilde{g}_{t},\tilde{x}_{t}-x^{*}\right\rangle .
\end{eqnarray*}
where we have defined $\tilde{g}_{t}:=\nabla f_{t}\left(\tilde{x}_{t}\right)$.
The high-level idea is to show that $R\left[X\right]\le o\left(T\right)$, which
implies $\mathbb{E}_{t}\left[f_{t}\left(\tilde{x}_{t}\right)-f_{t}\left(x^{*}\right)\right]\rightarrow0$
and thus convergence. First, we shall say something about each term
$\left\langle \tilde{g}_{t},\tilde{x}_{t}-x^{*}\right\rangle $.

\paragraph{Lemma 2:}

If $X=\mathbb{R}^{n}$, then for all $x^{*}$,
\begin{eqnarray*}
\left\langle \tilde{x}_{t}-x^{*},\tilde{g}_{t}\right\rangle  & = & \frac{1}{2}\eta_{t}\left\Vert \tilde{g}_{t}\right\Vert ^{2}+\frac{D\left(x^{*}\Vert x_{t}\right)-D\left(x^{*}\Vert x_{t+1}\right)}{\eta_{t}}
+ \left\langle \Delta_t, \tilde{g}_{t} \right\rangle
\end{eqnarray*}

\textbf{Proof:}

\begin{eqnarray*}
D\left(x^{*}\Vert x_{t+1}\right)-D\left(x^{*}\Vert x_{t}\right) & = & \frac{1}{2}\left\Vert x^{*}-x_{t}+x_{t}-x_{t+1}\right\Vert ^{2}-\frac{1}{2}\left\Vert x^{*}-x_{t}\right\Vert ^{2}\\
 & = & \frac{1}{2}\left\Vert x^{*}-x_{t}+\eta_{t}\tilde{g}_{t}\right\Vert ^{2}-\frac{1}{2}\left\Vert x^{*}-x_{t}\right\Vert ^{2}\\
 & = & \frac{1}{2}\eta_{t}^{2}\left\Vert \tilde{g}_{t}\right\Vert ^{2}-\eta_{t}\left\langle x_{t}-x^{*},\tilde{g}_{t}\right\rangle \\
 & = & \frac{1}{2}\eta_{t}^{2}\left\Vert \tilde{g}_{t}\right\Vert ^{2}-\eta_{t}\left\langle \tilde{x}_{t}-x^{*},\tilde{g}_{t}\right\rangle -\eta_{t}\left\langle x_{t}-\tilde{x}_{t},\tilde{g}_{t}\right\rangle \\
 & = & \frac{1}{2}\eta_{t}^{2}\left\Vert \tilde{g}_{t}\right\Vert ^{2}-\eta_{t}\left\langle \tilde{x}_{t}-x^{*},\tilde{g}_{t}\right\rangle - \eta_t \left\langle -\Delta_t, \tilde{g}_{t} \right\rangle
\end{eqnarray*}
Thus,
\begin{eqnarray*}
D\left(x^{*}\Vert x_{t+1}\right)-D\left(x^{*}\Vert x_{t}\right) & = & \frac{1}{2}\eta_{t}^{2}\left\Vert \tilde{g}_{t}\right\Vert ^{2}-\eta_{t}\left\langle \tilde{x}_{t}-x^{*},\tilde{g}_{t}\right\rangle - \eta_t \left\langle -\Delta_t, \tilde{g}_{t} \right\rangle\\
\frac{D\left(x^{*}\Vert x_{t+1}\right)-D\left(x^{*}\Vert x_{t}\right)}{\eta_{t}} & = & \frac{1}{2}\eta_{t}\left\Vert \tilde{g}_{t}\right\Vert ^{2}-\left\langle \tilde{x}_{t}-x^{*},\tilde{g}_{t}\right\rangle - \left\langle -\Delta_t, \tilde{g}_{t} \right\rangle\\
\left\langle \tilde{x}_{t}-x^{*},\tilde{g}_{t}\right\rangle  & = & \frac{1}{2}\eta_{t}\left\Vert \tilde{g}_{t}\right\Vert ^{2}+\frac{D\left(x^{*}\Vert x_{t}\right)-D\left(x^{*}\Vert x_{t+1}\right)}{\eta_{t}} + \left\langle \Delta_t, \tilde{g}_{t} \right\rangle.
\end{eqnarray*}
This completes the proof of Lemma 2. $\square$

\paragraph{Back to Theorem 1:}
Returning to the proof of Theorem 1, we use Lemma 2 to expand the regret $R[X]$:

\begin{eqnarray*}
R\left[X\right]\le\sum_{t=1}^{T}\left\langle \tilde{g}_{t},\tilde{x}_{t}-x^{*}\right\rangle  & = & \sum_{t=1}^{T}\frac{1}{2}\eta_{t}\left\Vert \tilde{g}_{t}\right\Vert ^{2}+\sum_{t=1}^{T}\frac{D\left(x^{*}\Vert x_{t}\right)-D\left(x^{*}\Vert x_{t+1}\right)}{\eta_{t}}
+\sum_{t=1}^{T} \left\langle \Delta_t, \tilde{g}_{t} \right\rangle\\
 & = & \sum_{t=1}^{T}\left[\frac{1}{2}\eta_{t}\left\Vert \tilde{g}_{t}\right\Vert ^{2}
 + \left\langle \Delta_t, \tilde{g}_{t} \right\rangle \right]\\
 &  & +\frac{D\left(x^{*}\Vert x_{1}\right)}{\eta_{1}}-\frac{D\left(x^{*}\Vert x_{T+1}\right)}{\eta_{T}}+\sum_{t=2}^{T}\left[D\left(x^{*}\Vert x_{t}\right)\left(\frac{1}{\eta_{t}}-\frac{1}{\eta_{t-1}}\right)\right]
\end{eqnarray*}
We now upper-bound each of the terms:
\begin{eqnarray*}
\sum_{t=1}^{T}\frac{1}{2}\eta_{t}\left\Vert \tilde{g}_{t}\right\Vert ^{2} & \le & \sum_{t=1}^{T}\frac{1}{2}\eta_{t}L^{2}\qquad\text{(Lipschitz assumption)}\\
 & = & \sum_{t=1}^{T}\frac{1}{2}\frac{\sigma}{\sqrt{t}}L^{2}\\
 & \le & \sigma L^{2}\sqrt{T},
\end{eqnarray*}
and
\begin{eqnarray*}
 &  & \frac{D\left(x^{*}\Vert x_{1}\right)}{\eta_{1}}-\frac{D\left(x^{*}\Vert x_{T+1}\right)}{\eta_{T}}+\sum_{t=2}^{T}\left[D\left(x^{*}\Vert x_{t}\right)\left(\frac{1}{\eta_{t}}-\frac{1}{\eta_{t-1}}\right)\right]\\
 & \le & \frac{F^{2}}{\sigma}+0+\frac{F^{2}}{\sigma}\sum_{t=2}^{T}\left[\sqrt{t}-\sqrt{t-1}\right]\qquad\text{(Bounded diameter)}\\
 & = & \frac{F^{2}}{\sigma}+\frac{F^{2}}{\sigma}\left[\sqrt{T}-1\right]\\
 & = & \frac{F^{2}}{\sigma}\sqrt{T},
\end{eqnarray*}
and
\begin{eqnarray*}
 &  & \sum_{t=1}^{T}\left\langle \Delta_t, \tilde{g}_{t} \right\rangle\\
 & \le & \left[\sum_{t=1}^{T} 2 v_t \sqrt{K}(P-1)L\right] \qquad\text{(Lemma 1 plus Lipschitz assumption)}\\
 & = & \left[\sum_{t=1}^{T} 2 \frac{\delta}{\sqrt{t}} \sqrt{K}(P-1)L\right]\\
 & \le & 4 \delta LP\sqrt{KT}.
\end{eqnarray*}
Hence,
\begin{eqnarray*}
R\left[X\right]\le\sum_{t=1}^{T}\left\langle \tilde{g}_{t},\tilde{x}_{t}-x^{*}\right\rangle  & \le & \sigma L^{2}\sqrt{T}+F^{2}\frac{\sqrt{T}}{\sigma} + 4 \delta LP\sqrt{KT}.
\end{eqnarray*}
This completes the proof of Theorem 1. $\square$

\section{Parameter Server Design and Implementation}
\label{sec:system-implementation}

\begin{figure}[th!]
\centering
  \includegraphics[width=0.9\textwidth]{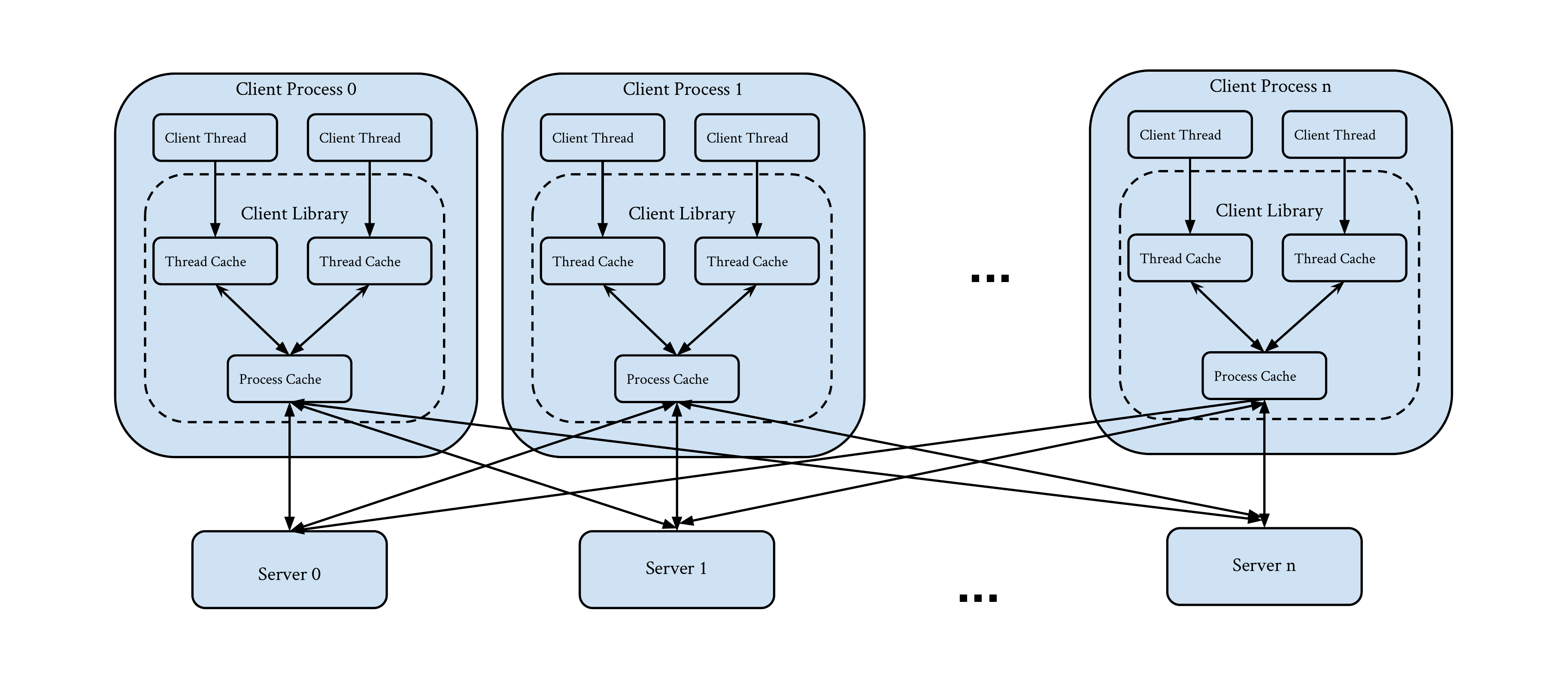}
  \caption{Parameter server architecture: a hierachical cache is used to minimize network communication overhead and reduce contention among application threads.}
    \label{fig:ps_arch}
\end{figure}

We implemented the proposed consitency models in a parameter server, called Petuum PS. 
For the purpose of comparison, Petuum PS also implemented SSP. Petuum PS is implemented in C++ and
ZeroMQ is used for network communication.

Petuum PS contains a collection of server processes, which holds the shared parameters in a distributed fashion..
An application process accesses the shared parameters via a client library. The client library caches the parameters in order
to minimize network communication overhead. An application process may contain multiple threads. The client library allocates a thread cache for each application thread to reduce contention among them. A thread is considered as a worker in our consistency mode description. The parameter server architecture is visualized in Fig~\ref{fig:ps_arch}. 

\subsection{System Abstraction}
Petuum PS organizes shared parameters as tables. Thus a parameter stored in Petuum PS is identified by a triple of table id, row id and column
id. Petuum PS supports both dense and sparse rows corresponding to dense and spares column index. A table is distributed across a set of server processes via hash partitioning and row is the unit of data distribution and transmission. The data stored in one table must be of the same type and Petuum PS can support an unlimited number of tables. It's worth mentioning that our implementation allows different tables to use different consistency model.

Petuum PS supports a set of straightforward APIs for data access. The main functions are:
\begin{itemize}
\item \texttt{Get(table\_id, row\_id, column\_id)}: Retrieve an element from a table. 
\item \texttt{Inc(table\_id, row\_id, column\_id, delta)}: Update an element by delta.
\item \texttt{Clock()}: Increment the worker thread's clock by 1.
\end{itemize}

\subsection{System Components}
Petuum PS's client library employs a two-level cache hierachy to hide network 
access latency and minimize contension among application threads: all 
application threads within a process share a \textbf{process cache} which caches
rows fetched from server. Each thread has its own \textbf{thread cache} and 
accesses to data are mostly serviced by thread cache as long as memory suffices.
In order to support asynchronous computation, thread cache employs a write-back
strategy.

In order to support Clock-based consistency models, the system needs to keep 
track of the clock of each worker. This is achieved via using vector clock.
Each client library maintains a vector clock where each entity represents the
clock of a thread. The minmum clock in the vector represents the progress of the 
process. Server treats a process as an entity and its vector clock keeps track
of the progress of all processes.

Asynchronous system tends to congest the network with large volume of messages. 
Our client and server thus batch messages to achieve high throughput. Messages 
are sent out based on their priorities which might be application-dependent. 
We by default prioritize updates with larger magnitude as they are more likely
to contribute to convergence.

\subsection{Implementing Consistency Models}

\begin{figure}
  \centering
  \includegraphics[width=0.5\textwidth]{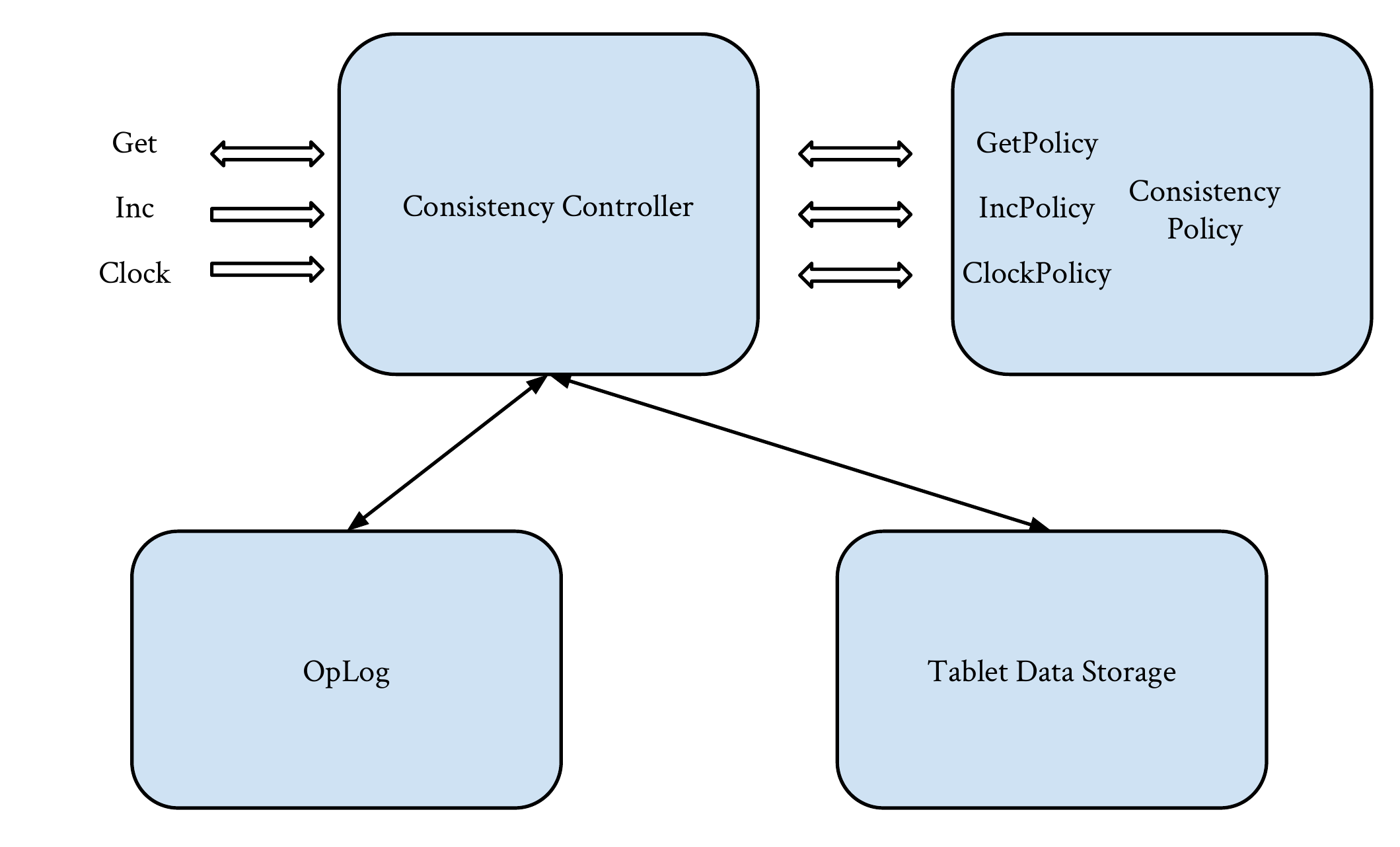}
  \caption{\small Consistency Controller and Consistency Policy}
  \label{fig:consistency_controller}
\end{figure}
 
We implement SSP, CAP, VAP and CVAP in a unified, modular fashion. We realized that those consistency models can be implemented by performing
different operations upon application threads accessing parameters. In other words, the exact sematics of the APIs depend on the consistency model being used. Each consistency model is expressed as a \emph{Consistency Policy} data structure. Each table is associated with a \emph{Consistency Controller}, which checks \emph{Consistency Policy} and services user accesses accordingly. The consistency control logic is visualized in Fig~\ref{fig:consistency_controller}.

Different semantics of the APIs include network communication and blocking wait 
for responses. Petuum PS uses three types of network communications:
\begin{itemize}
\item \textbf{Client Push}: Client pushes one or a batched set of updates to server. 
\item \textbf{Client Pull}: Client pull a row from server
\item \textbf{Server Push}: Server pushes one or a batched set of updates to relevant clients.
\end{itemize}

Coupled with proper cache coherence mechanism, those APIs are sufficient for 
implementing our consistency models.

\section{Evaluation}
\label{sec:evaluation}

We evaluate our prototype parameter server via topic modeling. Latent Dirichlet 
Allocation (LDA)~\cite{blei2003latent} is a popular unsupervised model that discovers a latent
topic vector for each document. We implemented LDA on our parameter server with the 
weak VAP model and conducted experiments on a 8-node cluster. Each node is equipped with
64 cores and 128GB of main memory and the nodes are connected with a 40 Gbps 
Ethernet network. We restrict our experiment to use at most 8 cores and 32GB of
memory per machine to emmulate cluster with normal machines.

We used a relatively small dataset 20News and evaluted the strong scalability of
the system in particular. Statistics of the 20News dataset are shown in 
Table~\ref{tbl:lda_corpra}.

\begin{table}
\begin{center}
  \begin{tabular}{ | c | c | }
    \hline
     & 20News \\ \hline
    \# of docs  & 11269 \\ \hline
    \# of words & 53485 \\ \hline
    \# of tokens& 1318299 \\ \hline
  \end{tabular}
  \caption{Summary statistics of two corpra used in LDA.}
  \label{tbl:lda_corpra}
\end{center}
\end{table}

\begin{figure}[th!]
\centering
  \includegraphics[width=1.5\textwidth]{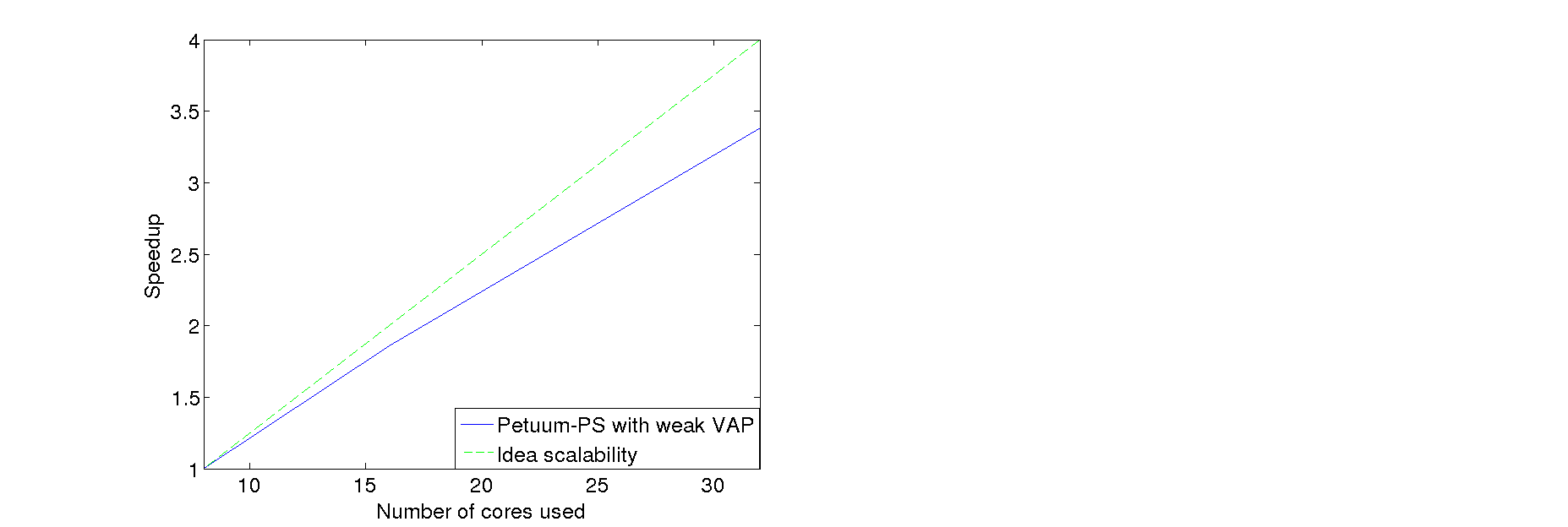}
    \label{fig:vap_scal}
\end{figure}

We fixed the number of topcis to be 2000 while varying the number of workers. 
We assign each worker a core exclusively. We show the results in 
Figure~\ref{fig:vap_scal} where the number of cores used ranges from 8 to 32.
The curve showed the speed up using Petuum-PS vs. ideal linear scalability.
Even though our experiments are conducted on a relatively small scale, the
results show that it has a great potential to scale up.

\subsubsection*{Acknowledgments}

We thank PRObE~\cite{Gibson+:login13} and CMU PDL Consortium for providing 
testbed and technical support for our experiments.

\bibliographystyle{plain}
\bibliography{ref}

\end{document}